\documentclass{article}
\usepackage{xcolor}
\usepackage{comment}
\usepackage{babel,blindtext}
\usepackage{graphicx} % Required for inserting images
\usepackage[normalem]{ulem}
\usepackage[square,numbers]{natbib}
\usepackage{subcaption}
\usepackage{hyperref}
\hypersetup{
    colorlinks=true,
    linkcolor=red,
    filecolor=magenta,
    urlcolor=blue,
    citecolor=purple,
    pdftitle={Overleaf Example},
    pdfpagemode=FullScreen,
    }
\usepackage{amsmath,amsfonts,amssymb}

\title{ Basis Vector Metric: A Method for Robust Open-Ended State Change Detection}
\author{David Oprea, Sam Powers}
\date{August 25, 2023}

\begin{document}
\maketitle

\begin{center}
Institutional Affiliation: Lumiere Foundation
\end{center}

\textbf{\textit{Abstract} - We test a new method, which we will abbreviate using the acronym BVM (Basis Vectors Method), in its ability to judge the state changes in images through using language embeddings. We used the MIT-States dataset, containing about 53,000 images, to gather all of our data, which has 225 nouns and 115 adjectives, with each noun having about 9 different adjectives, forming approximately 1000 noun-adjective pairs. For our first experiment, we test our method's ability to determine the state of each noun class separately against other metrics for comparison.  These metrics are cosine similarity, dot product, product quantization, binary index, Naive Bayes, and a custom neural network. Among these metrics, we found that our proposed BVM performs the best in classifying the states for each noun. We then perform a second experiment where we try using BVM to determine if it can differentiate adjectives from one another for each adjective separately. We compared the abilities of BVM to differentiate adjectives against the proposed method the MIT-States paper suggests: using a logistic regression model. In the end, we did not find conclusive evidence that our BVM metric could perform better than the logistic regression model at discerning adjectives. Yet, we were able to find evidence for possible improvements to our method; this leads to the chance of increasing our method's accuracy through certain changes in our methodologies.}

\section{Introduction}

Image classification has been a very widely studied subject in AI, with many methods nowadays used to perform multiple tasks that have appeared in this area of research. Although image classification has been widely researched, it mostly has been focused on static image classification. Static image classification is where you are just given an image and are tasked with determining what type of image it is. Yet, not much research focuses on dynamic image changes, where an image's properties can change and the model has the job of classifying what has changed for a certain image. Because of no research in dynamic image classification, tasks involving this are painstakingly hard to implement and are very expensive in resources, both computationally and time-wise. 

Usually, most image classification tasks involve just identifying an object. But for this study, we wanted to not focus on looking on what the type of object we are looking at is, but instead focus on the current "state" of the object. By using a dataset with static images, but with those images having different states, we can effectively simplify the problem, whereas one would likely use a video instead of images for this kind of task. From this, it is much easier to obtain data and convert the data into embeddings, which is the type of data we will use to evaluate all of our metrics. 

The main objective of this study is to help further our understanding of how to efficiently solve the task that is image state detection. We not only evaluate already made metrics and test them, but we also create a new, robust method, BVM. We created this new metric with the goal in mind of being able to discern subtle differences between images and using that to effectively classify the states of those images. We will go into detail about how we use supervised learning to try to make the important attributes of our images get more amplified and trivial ones diminished using our method. Also, the methods we present don't involve a lot of computing power or time. The creation of the embeddings from images is a pretty simple process, with many available models able to do this today each with its strengths. With a decent computer, you could easily utilize and/or quickly implement these metrics with thousands if not hundreds of thousands of image embeddings and obtain results from them. Also, some of the metrics we present don't require any pretraining at all, like cosine similarity or dot product. This means that once you have the embeddings for the images you want, all you have to do is quickly run the metric(s) through them and obtain your results. This gives our work the ability to be used by anyone to accomplish whatever task they need to do and whether they want to opt for speed or accuracy. This makes our research very easily usable and helpful in progressing the advance of dynamic image classification. 

\section{Related Works}

Though there is not much work in the field of dynamic image classification, there are countless works which helped us to become able to study this. 

\cite{imagenet} created a large image dataset used for improving the task of static image classification called ImageNet. This kickstarted the quick improvement of models in being able to classify images effectively into categories and it was also the inspiration for future image classifying competitions as well. \cite{clip-pae} performed a research project which was focused on taking an image and using some text to change the state of that image. The paper addresses this problem by introducing CLIP projection-augmentation embedding (PAE) to improve text-guided image manipulation through embeddings and residual vectors.R. Staniūtė et. al\cite{embedding_models} studied the performance of different embedding models on the Clickr30k and MSCOCO\cite{MSCOCO} datasets. It talks about AlexNet\cite{AlexNet}, ResNet\cite{Resnet}, VGGNet\cite{VGGNet}, and GoogLeNet (or Inception-X Net), making comparisons on how they performed. The researchers saw as the year of creation of the embedding models increased, they became more and more accurate at creating embeddings to describe images and their attributes. P. Sitikhu et. al\cite{cos-sim} researched testing the effectiveness of cosine similarity in the AG’s news topic classification data set.  They open up questions both on what is the right combination of embedding model and scoring method to use. 

\section{Methods}
The core concept underlying BVM\cite{Basis-Vectors} is that we want the basis vectors to be trained so that they exemplify the differences between two images to more accurately determine what their states are. We do this by making a matrix that has values of either 0 or 1, in which 0 will make values that we don't want to have much weight diminished, and 1 will try to increase the weight of the important values which easily differentiate images. This will in turn hopefully leave us with basis vectors that exemplify the state of the image through its values. 

To better visualize the idea of how basis vectors work we can look at how they manipulate vectors on a two-dimensional plane.\footnote{We generated the 2D plots for BVM on this Google Colab: \url{https://colab.research.google.com/drive/1efOLA3Fy_ht4pbFn3nnGiLjWvn5U43Qj##scrollTo=gEz-TJfHW2QH}.}

\begin{figure}[ht]
\centering
\begin{subcaptionblock}{0.3\textwidth}
  \includegraphics[scale = 0.4]{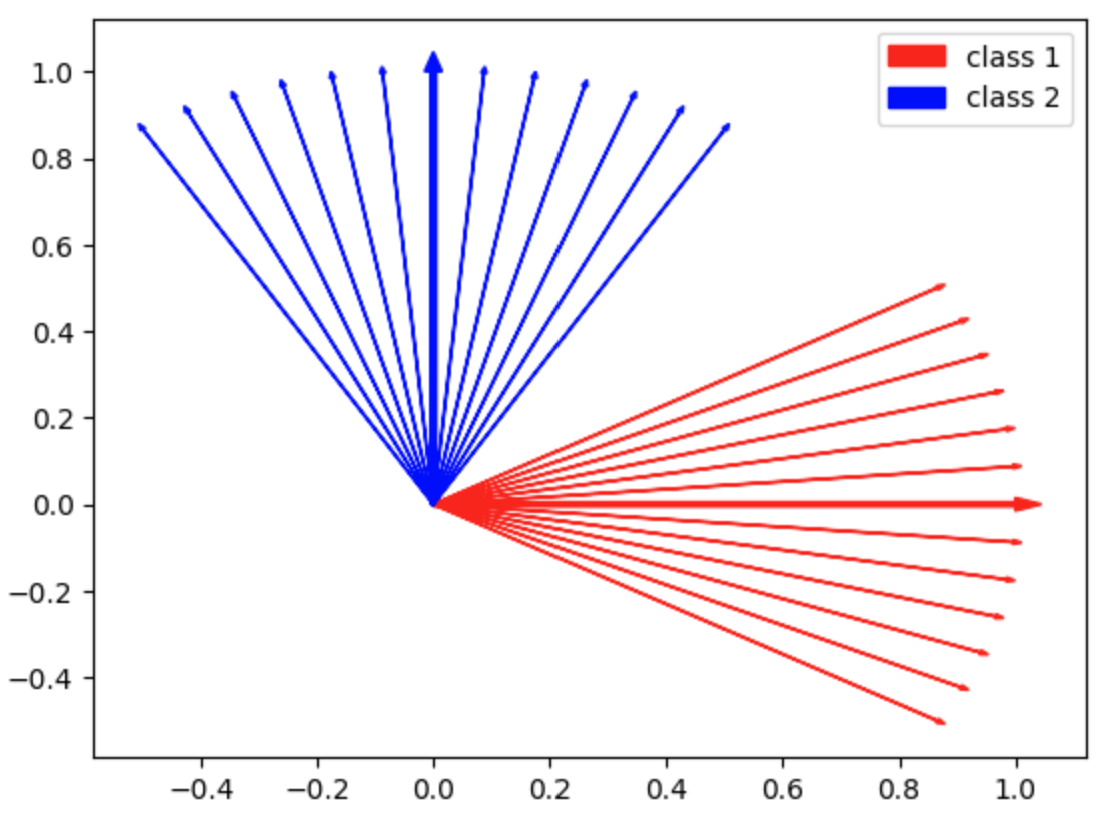}
  \caption{Initial Basis Vectors - Orthogonal Classes}
  \label{fig:basis_vectors_plot1}
\end{subcaptionblock}
\hspace{10em}
\begin{subcaptionblock}{0.3\textwidth}
  \includegraphics[scale = 0.4]{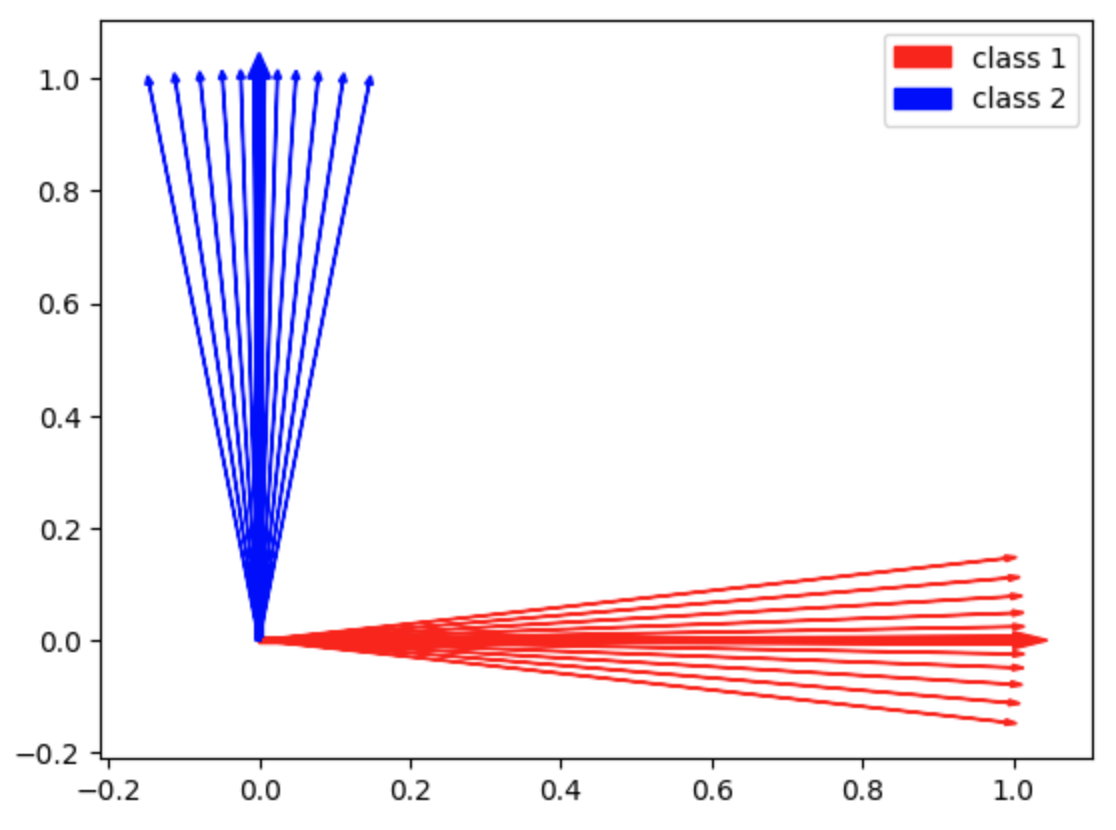}
  \caption{Trained Basis Vectors - Orthogonal Classes}
  \label{fig:basis_vectors_plot2}
\end{subcaptionblock}
\label{fig:basisvectors1}
\end{figure}

\begin{figure}[ht]
\centering
\begin{subcaptionblock}{0.3\textwidth}
  \includegraphics[scale = 0.4]{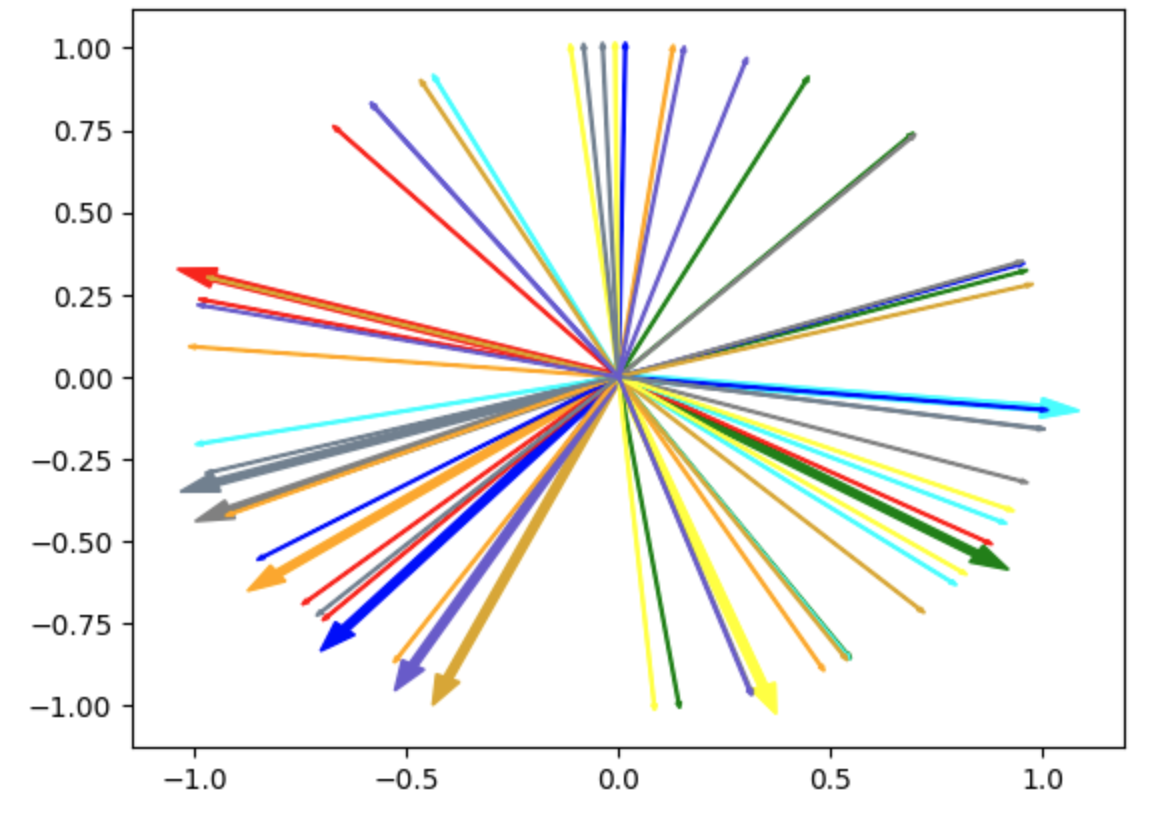}
  \caption{Initial Basis Vectors - 10 Random Classes}
  \label{fig:basis_vectors_plot3}
\end{subcaptionblock}
\hspace{10em}
\begin{subcaptionblock}{0.3\textwidth}
  \includegraphics[scale = 0.4]{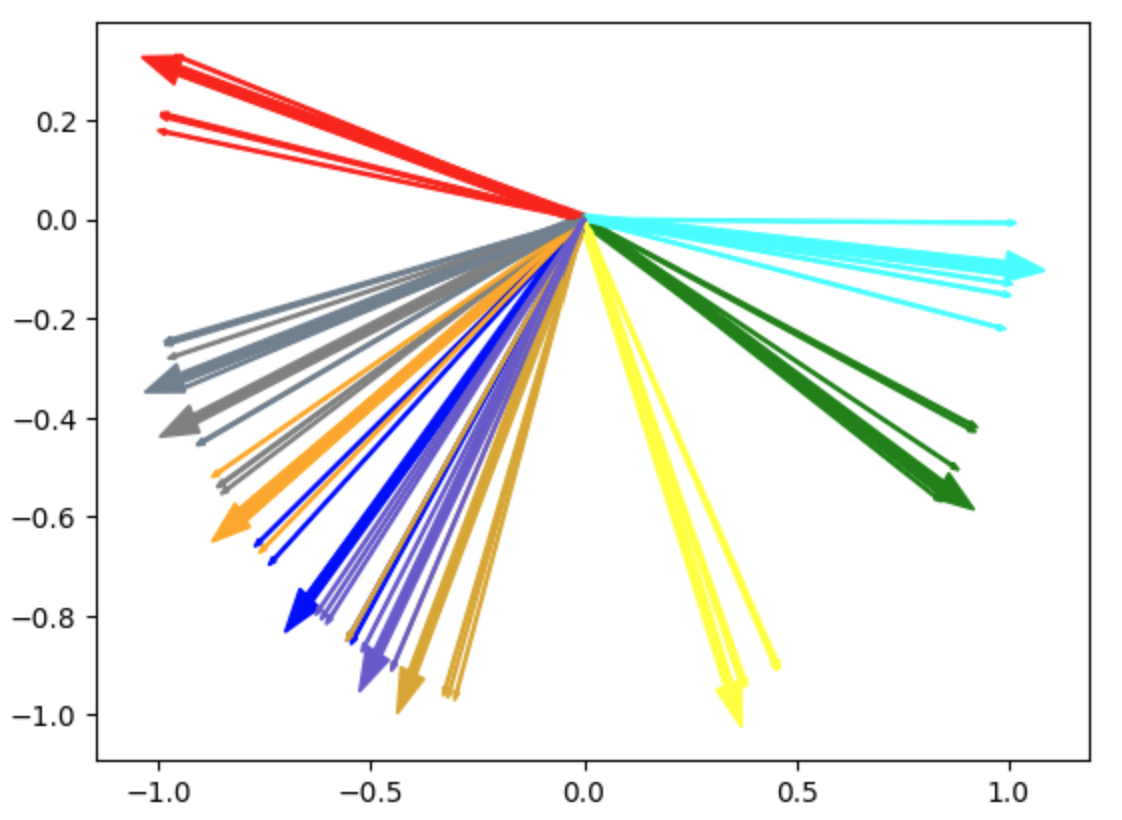}
  \caption{Trained Basis Vectors - 10 Random Classes}
  \label{fig:basis_vectors_plot4}
\end{subcaptionblock}
\label{fig:basisvectors2}
\end{figure}

In figures \ref{fig:basis_vectors_plot1} and  \ref{fig:basis_vectors_plot2} we see an example with two orthogonal classes, class 1 and class 2. Figure  \ref{fig:basis_vectors_plot1} shows untrained basis vectors to represent the classes going 5, 10, 15, 20, 25, and 30 degrees in both directions, with larger values showing more "unalike" vectors. Figure \ref{fig:basis_vectors_plot2} shows these basis vectors after being trained; it is clear to see that they move closer away towards their given class, and in general further away from their "opposite" class. Another example of this is shown in figures \ref{fig:basis_vectors_plot3}  and  \ref{fig:basis_vectors_plot4} where there are now ten classes. Figure \ref{fig:basis_vectors_plot3} shows the untrained basis vectors which are at first randomly rotated. Figure \ref{fig:basis_vectors_plot4} shows the basis vectors after training and again one can notice that the trained basis vectors are much closer to their initially given classes, better representing them. 

These visuals were using vectors that were only two-dimensional, but we can also see how basis vectors would act at any \textit{N}-dimensional scale using a TSNE plot.\footnote{We generated the TSNE plot for BVM on this Google Colab: \url{https://colab.research.google.com/drive/1t0L5rJZpCo8BocbeVPJo__oehBZQ_2uB?usp=sharing##scrollTo=aFprXjRhVSBt}. }

\begin{figure}
    \centering
    \includegraphics[scale = 0.4]{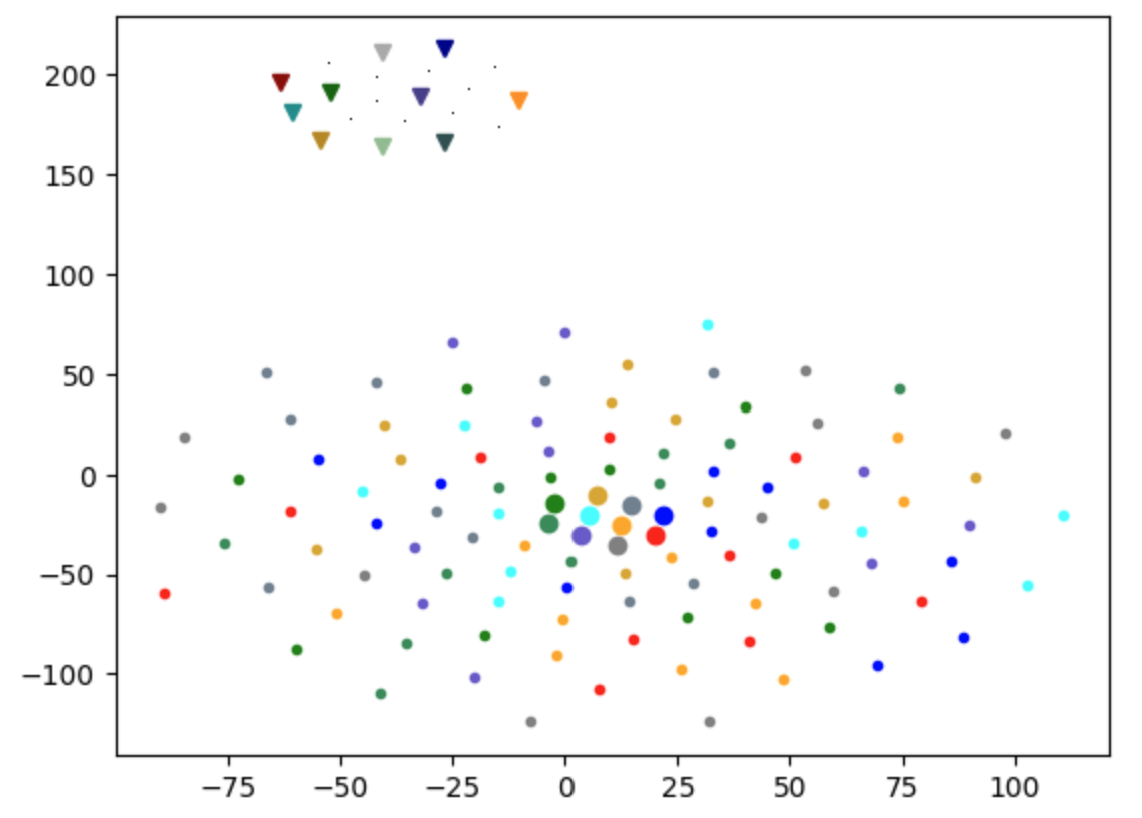}
    \caption{Basis Vectors operating on 100 dimensions}
    \label{fig:basis_vectors_plot5}
\end{figure}

Our TSNE plot on figure \ref{fig:basis_vectors_plot5} displays the initial basis vectors as tiny black dots. The trained basis vectors are represented with triangles, with their colors showing the class they are representing. The large colored dots are the true class values for each class, with their colors representing what class they belong to. Finally, the medium-sized dots represent the dataset that the basis vectors learned on, with their colors representing their true class. In this example, our basis vectors method scored a 95.96\% accuracy at identifying the correct class . 

Now that we intuitively understand BVM, here's a summary of how we implemented this method to be able to evaluate it: Firstly, we define three matrices: ${D}$ is the dataset with the embeddings used for training, ${B}$  is the basis vectors that are trained to accurately predict the state of an image, and ${T}$ is going to be your target matrix, with values of 0 and 1 that your basis vectors will try to match up with as closely as it can. We will also define a couple of more values: let $k$ be the number of states (adjectives) that your dataset contains, let $N$ be the number of training images in each class, and let $d$ be the dimensionality of the embeddings in your dataset (in this paper, our embeddings have a dimensionality of 768). ${D}$ will have dimensionality ($k*N$, $d$), ${B}$ will have dimensionality ($N$, $d$), and  ${T}$ will have dimensionality ($k$ * $N$, $k$). The $N$ vectors for the basis vectors ($B$) are each supposed to independently represent their adjective. To fill in the vectors for the basis vectors initially, we just took the averages of the embeddings for each adjective and used that value to represent the basis vector of that adjective. 

Afterward, we start to train the basis vectors using this equation for loss: 
$$
loss = \frac{1}{N} \sum{D \cdot B^\intercal - T}
$$

With the goal being:
$$
loss \rightarrow 0
$$

We first normalize our basis vectors to make sure they're no errors and the get the loss. After this we import the Adam optimizer from Pytorch, back-propagating the loss, and step the optimizer. We repeat this for $x$ amount of epochs. 

Then to get the match scores  (${M}$) for a query image you can use the query image's embedding (${Q}$) and plug in the equation: 

$$
M = {Q \cdot B^\intercal \cdot T^\intercal}
$$

Finally to find out which adjective the basis vector matched to the most similar to, just find what column in match scores has the largest value. 

\subsection*{\textit{Image Embeddings}}

Image embeddings are a vector representation of an image obtained through a neural network trained on a large dataset. These vectors contain numbers that are supposed to represent the contents of the image. The purpose of this is to simplify the data of the image and to make models more easily compare different images. 

To obtain our main image embeddings we decided to use the CLIP-ViT-Large-Patch14 embedding model\cite{CLIP} provided by the Transformers module from HuggingFace. We decided this since it produces accurate embeddings and the embeddings have small dimensionality, only having 768 numbers. With the small dimensionality, the creation of these embeddings is quick and easy. 

\section{Experimental Setup}

\subsection*{\textit{Dataset}}

In this study to obtain our results we used the MIT-States dataset\cite{MIT-States}. It is a dataset that contains images that are classified by both an adjective and a noun. From this, we can use the adjectives of images as their "states" that we can use to evaluate the results of our metrics. This dataset contains images of 245 nouns and 115 adjectives, with each noun having about 9 adjective varieties. 

\begin{figure}[ht]
\centering
\begin{subcaptionblock}{0.3\textwidth}
  \includegraphics[width=\linewidth]{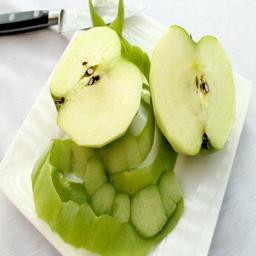}
  \caption{"Peeled Apples"}
  \label{fig:peeled_apples}
\end{subcaptionblock}
\hspace{1em}
\begin{subcaptionblock}{0.3\textwidth}
  \includegraphics[width=\linewidth]{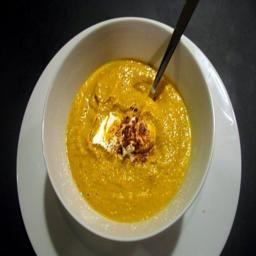}
  \caption{"Pureed Apples"}
  \label{fig:pureed_apples}
\end{subcaptionblock}
\\
\begin{subcaptionblock}{0.3\textwidth}
  \includegraphics[width=\linewidth]{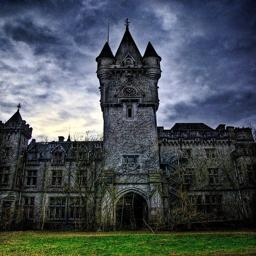}
  \caption{"Ancient Castle"}
  \label{fig:ancient_castle}
\end{subcaptionblock}
\hspace{1em}
\begin{subcaptionblock}{0.3\textwidth}
  \includegraphics[width=\linewidth]{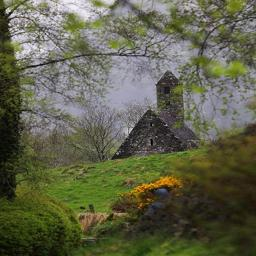}
  \caption{"Ancient Church"}
  \label{fig:ancient_church}
\end{subcaptionblock}

\caption{Example images from the dataset. (a) and (b) show different adjectives for the same noun, and (c) and (d) show different nouns for the same adjective.}
\label{fig:dataset_ex}
\end{figure}

A couple example images from this dataset can be found on figure \ref{fig:dataset_ex}

\subsection*{\textit{Baseline Metrics}}

\begin{figure}
    \centering
    \includegraphics[scale = 0.35]{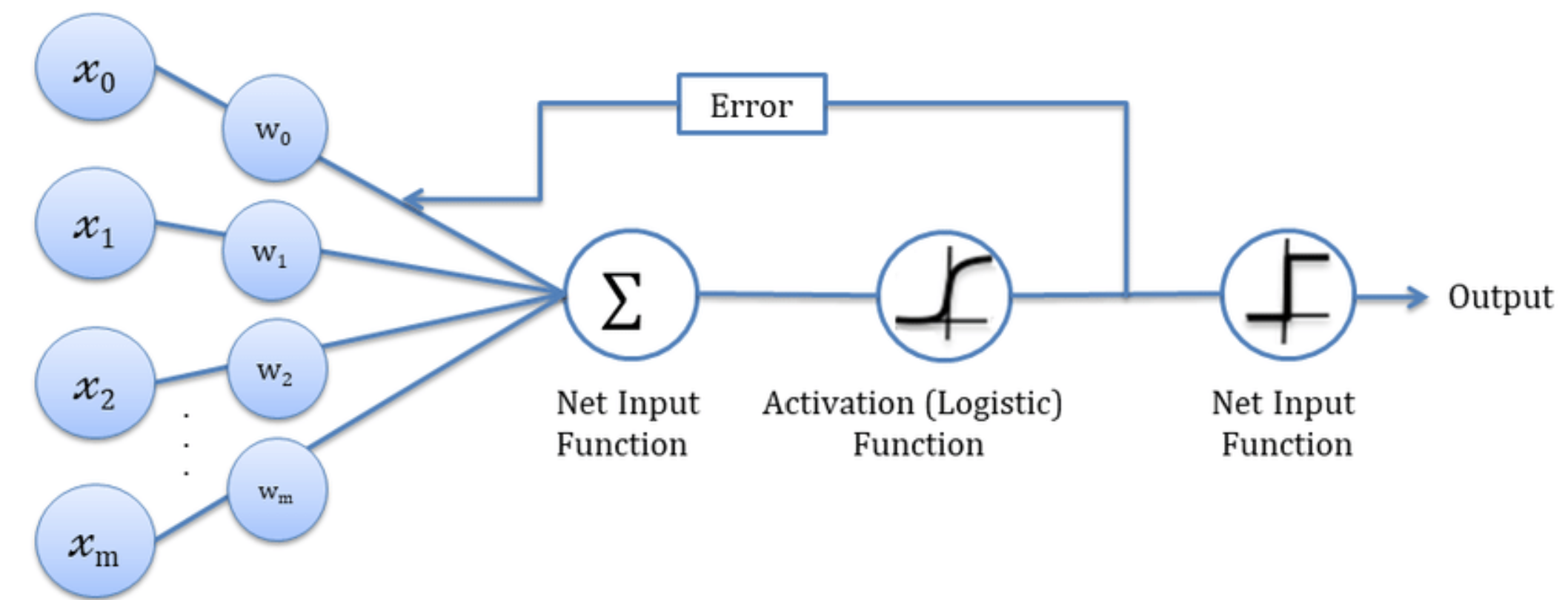}
    \caption{A diagram of how our logistic regression model would work for discerning different adjectives \cite{logisticregression}}
    \label{fig:logisticregression}
\end{figure}

For out study, we compared to seven baseline metrics. We implemented these metrics using the Faiss library\cite{Faiss}, Pytorch\cite{Pytorch}, Numpy\cite{Numpy}, and Pandas: 

\begin{enumerate}
    \item Cosine Similarity: Out first baseline metric, this metric measures the similarity between two vectors of an inner product space. It outputs a value from 0 to 1, with 0 meaning the vectors are the same and 1 meaning they are different. It gets this value by using the equation: 

$$
CS = \frac{A \cdot B}{||A|| \cdot ||B||}
$$
Where $CS$ is the cosine similarity value, $A$ is the first matrix, $B$ is the second matrix, and $||X||$ is the magnitude of matrix $X$. 
    
    \item Dot Product: Our second baseline metric, this metric measures the similarity between two vectors through the sum of the products of their elements with regards to position: 

$$
DP = \sum_{i=0}^{N-1}{A_i \cdot B_i}
$$

Where $DP$ is the dot product value, $N$ is the length of the vectors (they should be the same length), $A_i$ is the \textit{i}-th index of vector $A$, and $B_i$ is the \textit{i}-th index of vector $B$. The greater the value it outputs, the greater the similarity between the two vectors. 
    
    \item Binary Index: This metric stores the vectors as arrays of bytes, so that a vector of size \textit{d} takes only \textit{d}/8 bytes in memory. The metric then computes the hamming distances of these converted vectors to each other: 

$$
HD = \frac{1}{N}\sum_{i=0}^{N-1}{A_i - B_i}
$$

Where $HD$ is the hamming distance value, $N$ is the length of the vectors (they should be the same length), $A_i$ is the \textit{i}-th index of vector $A$, and $B_i$ is the \textit{i}-th index of vector $B$.  To make the embedding values valid for indexing, I normalized them to be integer values between 0 and 255, which Faiss recommends doing for effective indexing and good accuracy. 

\item Product Quantization: This metric keeps the number of vectors the same after quantization. However, the values in the compressed vector are now transformed into short codes, and therefore they are symbolic and are no longer numeric. The metric then computes the hamming distance to these two metrics, similar to the binary index metric I have mentioned previously. 
    
    \item Naive Bayes: A classical probabilistic classifier based on Bayes' theorem:

$$
P(Y=k|X_1,X_2...X_n) = \frac{P(Y)\prod_{i=0}^{N-1}{P(X_i|Y)}}{P(X_1)*P(X_2)...*P(X_n)}
$$

    We trained it on the training embeddings for each adjective-noun pair and then tested it on the testing embeddings to obtain our results. 
    
    \item Custom Neural Network: A very simple neural network that has a dense input layer of (20, 768), a hidden linear layer of (768, 5), and an output linear layer of (5, \# of adjectives for the noun class). You can see a diagram of it in figure \ref{fig:customneuralnetwork}

\begin{figure}
    \centering
    \includegraphics[scale = 0.3]{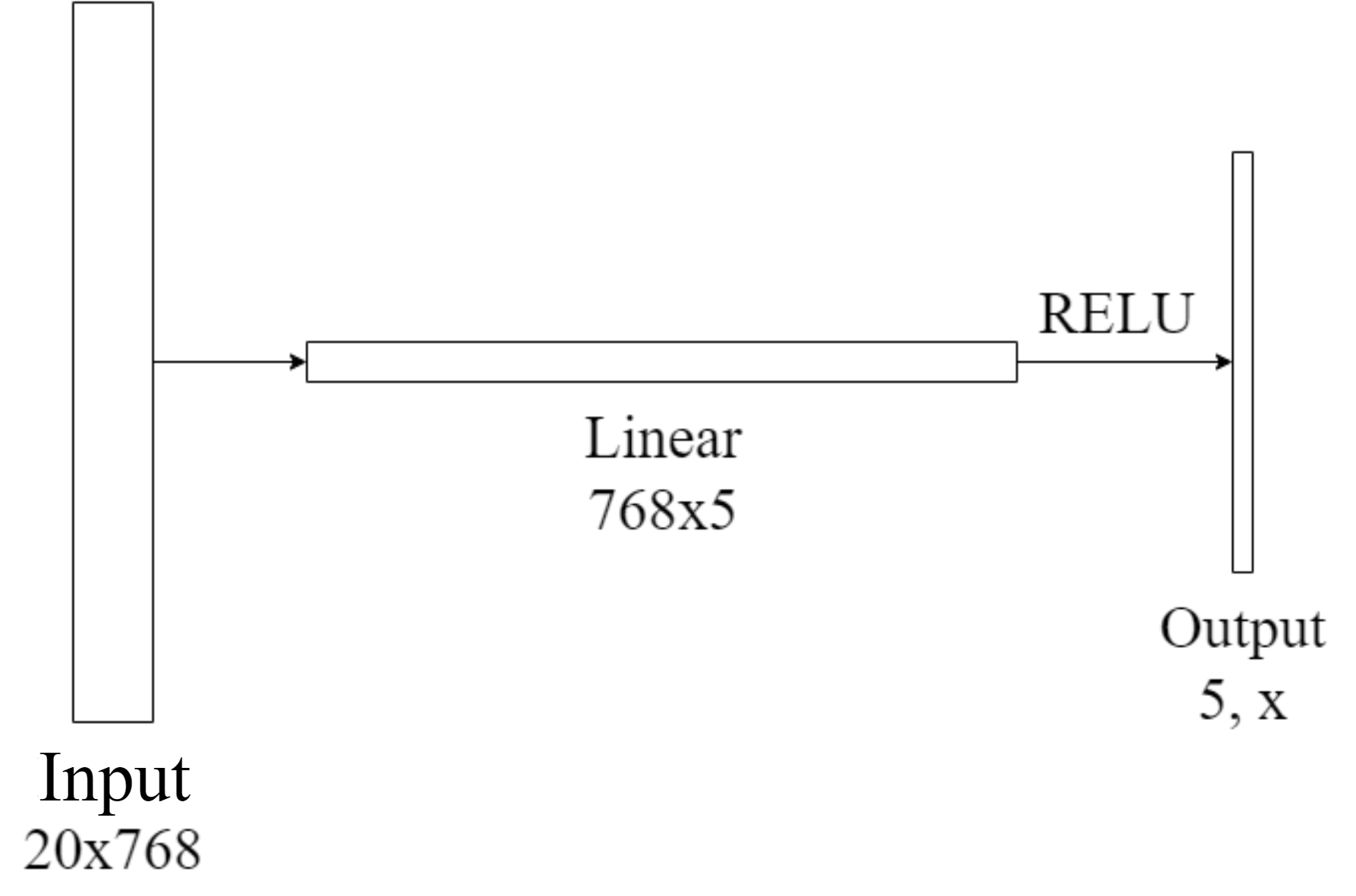}
    \caption{The basic neural network we created as a metric for comparison}
    \label{fig:customneuralnetwork}
\end{figure}
    
    \item Logistic Regression: Estimates the class of the input variable into either the 1st class or the 2nd class. This was only used for the adjectives section of the experiment. You can see a diagram of how our logistic regression model would work in figure \ref{fig:logisticregression}
\end{enumerate}

\section{Results}

\subsection*{\textit{Noun-Adjective Pairs Testing}}

For this test, we evaluated how accurately our metrics correctly identified the adjective (state) for each noun. 

We first organized the data to make sure metrics had enough images to utilize for training for each adjective-noun pairing. We did this by filtering out all folders containing 20 images or less for our noun.

We then created a map that would have an array for each noun class. Then we would further separate the noun classes through their adjectives. In each noun-adjective class, we would add the corresponding images to them. 

After having the organized map, we would run a for-each loop that would go through all the nouns. We would then have another loop going through the adjectives and getting the embeddings for all the images. The first 20 embeddings for each adjective class would be used for training, and the rest would be used for testing. Finally, we run the metrics and print out the results for each noun class to utilize. 

\begin{figure}[ht]
\centering
\begin{subcaptionblock}{0.45\textwidth}
  \includegraphics[width=\linewidth]{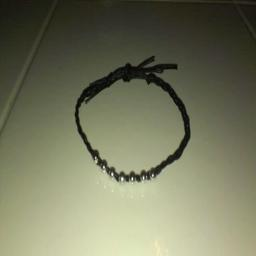}
  \caption{Training Image - True Value: Thin Cord\\
  Cosine Similarity Prediction: Thin Cord \\
  Dot Product Prediction: Thin Cord \\
  Product Quantization Prediction: Thin Cord\\
  Binary Index Prediction: Thin Cord\\
  Naive Bayes Prediction: Thin Cord \\
  Basis Vectors Prediction: Thin Cord\\
  Custom Neural Network Prediction: Thin Cord}
  \label{fig:traincord}
\end{subcaptionblock}
\hspace{1em}
\begin{subcaptionblock}{0.45\textwidth}
  \includegraphics[width=\linewidth]{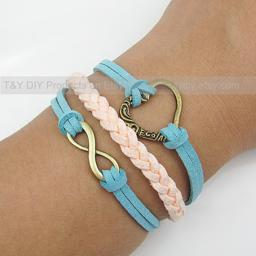}
  \caption{Training Image - True Value: Thin Cord\\
  Cosine Similarity Prediction: Thin Cord,\\
  Dot Product Prediction: Thin Cord \\
  Product Quantization Prediction: Thin Cord \\
  Binary Index Prediction: Thin Cord\\
  Naive Bayes Prediction: Thin Cord \\
  Basis Vectors Prediction: Thin Cord \\
  Custom Neural Network Prediction: Thin Cord\\}
  \label{fig:testcord}
\end{subcaptionblock}
\\
\begin{subcaptionblock}{0.45\textwidth}
  \includegraphics[width=\linewidth]{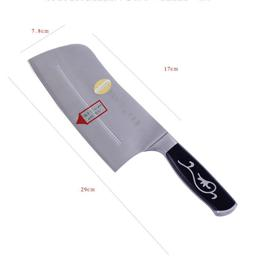}
  \caption{Training Image -
True Value: Sharp Blade\\
Cosine Similarity Prediction: Narrow Blade\\
Dot Product Prediction: Narrow Blade\\
Product Quantizer Prediction: Narrow Blade\\
Binary Index Prediction: Narrow Blade\\
Naive Bayes Prediction: Bent Blade\\
Basis Vectors Prediction: Wide Blade\\
Custom Neural Network Prediction: Curved Blade\\}
  \label{fig:trainblade}
\end{subcaptionblock}
\hspace{1em}
\begin{subcaptionblock}{0.45\textwidth}
  \includegraphics[width=\linewidth]{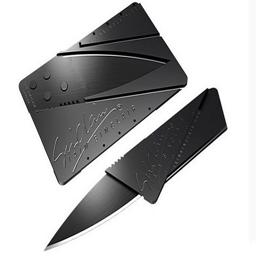}
  \caption{Training Image -
True Value: Sharp Blade\\
Cosine Similarity Prediction: Narrow Blade\\
Dot Product Prediction: Narrow Blade\\
Product Quantizer Prediction: Narrow Blade\\
Binary Index Prediction: Narrow Blade\\
Naive Bayes Prediction: Bent Blade\\
Basis Vectors Prediction: Wide Blade\\
Custom Neural Network Prediction: Curved Blade\\}
  \label{fig:testblade}
\end{subcaptionblock}

\caption{Some results from noun adjective testing with both training and testing images. Both correct and incorrect predictions for all metrics are shown on here.}
\label{fig:NounAdjEx}
\end{figure}

Some examples of different images and the performances of metrics on these images are found on figure \ref{fig:NounAdjEx}

The average accuracies for each method are reported in Table~\ref{table:acc_noun_adj}.

\begin{table}
    \begin{tabular}{ |p{1.5cm}|p{1.5cm}|p{1.5cm}|p{1.5cm}|p{1.5cm}|p{1.5cm}|p{1.5cm}|}
     \hline
     \multicolumn{7}{|c|}{Averages for Metrics on Noun-Adjective Pairs} \\
     \hline
     Cosine Similarity & Dot Product  & Binary Index & Product Quantizor & Naive Bayes & Custom Neural Network & Basis Vectors\\
     \hline
    55.99\%&45.43\%&52.17\%&41.95\%&65.23\%&22.99\%& \textbf{66.14\%}\\
     \hline
    \end{tabular}

\caption{Average Accuracies of metrics and their performances on Noun-Adjective Pairs.}
\label{table:acc_noun_adj}
\end{table}

\subsection*{\textit{Adjectives Testing}}

After testing and comparing metrics and their ability to discern adjectives only for separate noun classes, we were interested in what would happen if we just tried discerning adjectives. 

We did this by putting images into adjective classes, again using a map. We then randomly shuffle the images in each class so that it is likely that both the training and testing sets would have an even distribution of noun classes. We then went through the images and inserted their embedding into the corresponding adjective classes. We then split the embeddings for each class by 80:20 (80\% in the training set and 20\% in the testing set). We then tested how well our models could tell apart each adjective from all the other adjectives. We did this by using a loop and then the selected adjective would be class 0 and the other adjectives would be class 1.\\

Here are the results we obtained: \\

\begin{center}
\begin{tabular}{ |p{3cm}|p{3cm}|}
 \hline
 \multicolumn{2}{|c|}{Averages for Metrics on Discerning Adjectives using CLIP-ViT-Large-Patch14} \\
 \hline
 Logistic Regression & Basis Vectors\\
 \hline
\textbf{45.13\%}& 40.46\%\\
\hline
\end{tabular}
\label{table:adj_test1}
\end{center}

\begin{center}
\begin{tabular}{ |p{3cm}|p{3cm}|}
 \hline
 \multicolumn{2}{|c|}{Averages for Metrics on Discerning Adjectives using VGG19} \\
 \hline
 Logistic Regression & Basis Vectors\\
 \hline
4.66\% & \textbf{4.71\%}\\
 \hline
\end{tabular}
\label{table:adj_test2}
\end{center}

Based on table \ref{table:adj_test1}, we found that logistic regression performed better than our BVM metric, even when it was trained on 5000 epochs. Though we could not get the results we wanted, we investigated further and found that using a different embedding model, VGG19, we found that our method was able to perform better than logistic regression as shown in table \ref{table:adj_test2}. Though the overall accuracies are worse, BVM did get a higher accuracy. This means that with the right embedding model, BVM could perform better than the logistic regression method that the MIT States paper proposed. 

\begin{figure}[ht]
\centering
\begin{subcaptionblock}{0.45\textwidth}
  \includegraphics[width=\linewidth]{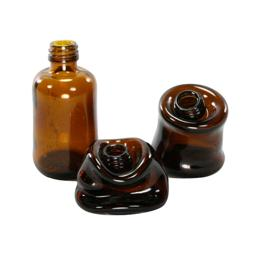}
  \caption{Training Image - True Value: Melted\\
  Logistic Regression Prediction: Melted \\
  BVM Prediction: Melted }
  \label{fig:truemelted}
\end{subcaptionblock}
\hspace{1em}
\begin{subcaptionblock}{0.45\textwidth}
  \includegraphics[width=\linewidth]{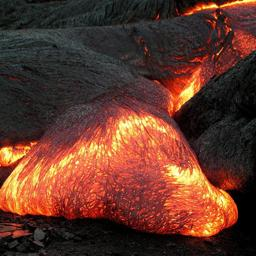}
\caption{Training Image - True Value: Melted\\
  Logistic Regression Prediction: Molten \\
  BVM Prediction: Molten }
  \label{fig:falsemelted}
\end{subcaptionblock}
\\
\begin{subcaptionblock}{0.45\textwidth}
  \includegraphics[width=\linewidth]{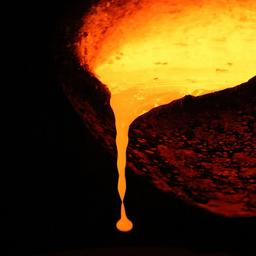}
 \caption{Testing Image - True Value: Molten\\
  Logistic Regression Prediction: Molten \\
  BVM Prediction: Molten }
  \label{fig:truemolten}
\end{subcaptionblock}
\hspace{1em}
\begin{subcaptionblock}{0.45\textwidth}
  \includegraphics[width=\linewidth]{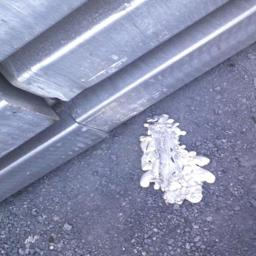}
\caption{Testing Image - True Value: Molten\\
  Logistic Regression Prediction: Melted \\
  BVM Prediction: Melted }
  \label{fig:falsemolten}
\end{subcaptionblock}

\caption{Some results from adjective testing with both training and testing images. Both correct and incorrect predictions for all metrics are shown on here.}
\label{fig:AdjEx}
\end{figure}

Some example results can be found on figure \ref{fig:AdjEx}.

\section{Discussion}
As we can see, BVM performed the best overall in the noun-adjective pairs category, being slightly better than Naive Bayes. Though this gap could widen further if we trained BVM for more than 1000 epochs, whereas we can not change anything about the implementation for our Naive Bayes model. 

Though BVM did perform worse than logistic regression for the adjective testing, we were able to determine that there is potential for improvement if we change our methodology for embedding creation, as we saw using the VGG19 embedding model. Also, the dataset we used contains some inconsistent images,  a clear example is shown in image \ref{fig:falsemelted}. If inconsistent images like this appear sparsely, it is reasonable to see how it could greatly throw off BVM's predictions.

\begin{comment}
We were able to improve the researchers from MIT-States' idea of using a logistic regressions model for differentiating adjectives through using basis vectors, coming very close to being able to discern one adjective from others in terms of accuracy.  

Overall, this confirms that our idea for basis vectors is possibly a good idea for a metric to use for dynamic image classification given that it performed the best out of our other metrics and that they have the potential to perform even better since we could've trained the vectors even more. 
\end{comment}

A limitation of this study was that we didn't have a large amount of time due to the deadlines of this paper needing to be met. Potential future research could include looking into using these metrics with different embedding models to see what type of encoder (embedding) model works the best with either the current premade metrics that we tested, or our robust method, BVM. Also, one could look into how one could fine-tune the training of our method so that inconsistent image attributes are ignored more. If BVM were to focus on reliable images, there is likely great promise for a much better overall accuracy. 

If these tasks were to be accomplished, BVM could likely be a very powerful metric in the task of dynamic image classification. 

\section{Conclusion}

We found that our robust method, BVM, performs better than our other metrics for the nouns-adjective pairs testing, but not for the adjective testing. Though, we know that with the right embedding model and/or fine-tuning for training, there is potential for BVM's accuracy to improve. We did both of our tests on the MIT States dataset and we created our main training and testing embeddings using the CLIP-ViT-Large-Patch14 model. 

Our method is compelling because it focuses on amplifying the attributes of embeddings that differentiate them, allowing our method to be more accurate in determining what class each embedding will be in. It's also very simple to implement and you can easily see how the method is learning whilst you train it. This in turn makes it a very fast process to debug and fine-tune our method. 

Also, the metrics that we tested don't require a lot of computing power to run on your computer, with the creation of the embeddings being the longest part of the whole process. 

These results let us know that BVM could potentially lead to the best outcome in results if one were to use a metric to solve a task that involved state changes of images. We hope that our work with both already established metrics and our robust metric, BVM, helps get rid of some of the troubles of helping researchers know the right metric to pick. We look forward to seeing this paper hopefully leading to an acceleration in the progress of trying to solve the dynamic image classification task.

\clearpage
\appendix
\section*{Appendices}
Our full results for the noun-adjective pairs experiment:

\begin{center}

\begin{tabular}{ |p{1.5cm}|p{1.5cm}|p{1.5cm}|p{1.5cm}|p{1.5cm}|p{1.5cm}|p{1.5cm}|p{1.5cm}|}
 \hline
 \multicolumn{8}{|c|}{Results for Metrics on Noun-Adjective Pairs} \\
 \hline
 Noun & Cosine Similarity & Dot Product & Binary Index & Product Quantiza- tion & Naive Bayes & Custom Neural Network & Basis Vectors\\
 \hline
 aluminum&61.75\%&53.4\%&62.8\%&44.52\%&\textbf{72.55}\%&12.5\%&71.57\%\\
animal&61.12\%&28.73\%&59.02\%&42.0\%&64.56\%&20.0\%&\textbf{72.15}\%\\
apple&64.03\%&22.93\%&41.38\%&26.06\%&61.17\%&14.29\%&\textbf{67.96}\%\\
armor&\textbf{100.0}\%&100.0\%&100.0\%&100.0\%&100.0\%&100.0\%&100.0\%\\
bag&47.93\%&43.92\%&50.38\%&33.27\%&\textbf{52.94}\%&8.33\%&52.94\%\\
ball&63.33\%&36.67\%&53.33\%&54.27\%&54.29\%&33.33\%&\textbf{80.0}\%\\
balloon&36.06\%&14.29\%&32.29\%&36.07\%&\textbf{61.4}\%&14.29\%&51.75\%\\
banana&60.37\%&29.48\%&54.39\%&40.81\%&\textbf{63.37}\%&16.67\%&61.39\%\\
basement&54.93\%&48.62\%&47.33\%&41.18\%&54.37\%&14.29\%&\textbf{55.34}\%\\
basket&75.04\%&49.84\%&78.53\%&72.66\%&84.69\%&20.0\%&\textbf{86.73}\%\\
bathroom&31.34\%&22.05\%&25.19\%&19.3\%&32.37\%&7.69\%&\textbf{35.27}\%\\
bay&76.03\%&60.29\%&66.08\%&50.5\%&81.25\%&25.0\%&\textbf{85.94}\%\\
beach&50.62\%&37.3\%&47.86\%&43.79\%&66.67\%&25.0\%&\textbf{71.6}\%\\
bean&\textbf{61.79}\%&44.7\%&57.44\%&38.63\%&56.34\%&20.0\%&53.52\%\\
bear&50.88\%&33.33\%&60.83\%&68.42\%&79.55\%&33.33\%&\textbf{81.82}\%\\
bed&76.22\%&77.41\%&75.68\%&71.37\%&89.29\%&40.62\%&\textbf{89.29}\%\\
beef&51.49\%&44.66\%&33.27\%&19.16\%&\textbf{59.79}\%&9.09\%&58.25\%\\
belt&44.92\%&46.75\%&43.67\%&38.75\%&\textbf{56.32}\%&20.0\%&55.17\%\\
berry&85.42\%&68.75\%&73.96\%&81.25\%&\textbf{92.86}\%&50.0\%&89.29\%\\
bike&52.33\%&51.45\%&55.54\%&40.48\%&\textbf{58.14}\%&16.67\%&55.81\%\\
blade&33.5\%&30.72\%&35.57\%&20.31\%&\textbf{49.44}\%&12.5\%&42.7\%\\
boat&52.69\%&46.91\%&54.45\%&48.99\%&\textbf{72.94}\%&14.29\%&72.94\%\\
book&64.45\%&29.13\%&64.38\%&49.87\%&\textbf{73.03}\%&16.67\%&69.66\%\\
bottle&55.66\%&46.35\%&48.71\%&29.47\%&62.63\%&9.09\%&\textbf{63.16}\%\\
boulder&49.21\%&44.05\%&44.77\%&40.49\%&56.38\%&16.67\%&\textbf{59.57}\%\\
bowl&61.09\%&57.24\%&59.53\%&49.79\%&\textbf{66.9}\%&14.29\%&66.21\%\\
box&48.33\%&25.94\%&45.2\%&37.38\%&\textbf{56.63}\%&16.67\%&54.22\%\\
bracelet&47.49\%&44.26\%&36.74\%&24.65\%&\textbf{56.1}\%&11.11\%&53.17\%\\
branch&47.32\%&42.63\%&55.81\%&30.0\%&\textbf{64.79}\%&16.67\%&63.38\%\\
brass&42.76\%&41.56\%&41.15\%&27.38\%&52.36\%&9.09\%&\textbf{52.79}\%\\
bread&40.2\%&21.61\%&32.93\%&26.72\%&\textbf{52.17}\%&10.0\%&51.09\%\\
bridge&44.56\%&42.18\%&51.62\%&43.55\%&\textbf{61.29}\%&18.06\%&60.22\%\\
bronze&65.16\%&60.26\%&71.22\%&39.33\%&71.32\%&12.5\%&\textbf{72.79}\%\\
bubble&72.73\%&50.0\%&\textbf{86.36}\%&81.82\%&77.27\%&50.0\%&86.36\%\\
bucket&60.0\%&66.67\%&40.0\%&46.67\%&75.0\%&40.0\%&\textbf{89.29}\%\\
building&59.81\%&45.54\%&52.58\%&31.42\%&66.52\%&10.0\%&\textbf{66.52}\%\\
 \hline
\end{tabular}

\begin{tabular}{ |p{1.5cm}|p{1.5cm}|p{1.5cm}|p{1.5cm}|p{1.5cm}|p{1.5cm}|p{1.5cm}|p{1.5cm}|}
 \hline
 \multicolumn{8}{|c|}{Results for Metrics on Noun-Adjective Pairs} \\
 \hline
 Noun & Cosine Similarity & Dot Product & Binary Index & Product Quantiza- tion & Naive Bayes & Custom Neural Network & Basis Vectors\\
 \hline
 bus&45.23\%&45.02\%&47.19\%&31.17\%&\textbf{68.67}\%&11.11\%&66.27\%\\
bush&74.62\%&64.62\%&\textbf{86.15}\%&72.31\%&66.67\%&52.31\%&66.67\%\\
butter&39.46\%&38.1\%&51.87\%&20.41\%&\textbf{52.54}\%&14.29\%&49.15\%\\
cabinet&44.78\%&41.4\%&43.67\%&37.37\%&45.75\%&12.5\%&\textbf{47.71}\%\\
cable&\textbf{95.24}\%&81.43\%&81.43\%&69.35\%&91.43\%&33.33\%&91.43\%\\
cake&62.71\%&56.65\%&53.26\%&40.8\%&\textbf{73.96}\%&20.0\%&67.71\%\\
camera&55.38\%&44.59\%&55.69\%&43.76\%&\textbf{70.68}\%&12.5\%&69.92\%\\
candle&63.33\%&35.56\%&55.0\%&47.94\%&75.0\%&33.33\%&\textbf{79.55}\%\\
candy&62.93\%&34.6\%&41.92\%&45.73\%&78.72\%&25.0\%&\textbf{87.23}\%\\
canyon&49.09\%&39.86\%&59.39\%&33.15\%&59.63\%&14.29\%&\textbf{67.89}\%\\
car&65.92\%&54.1\%&61.52\%&45.86\%&\textbf{67.62}\%&11.11\%&66.19\%\\
card&53.27\%&57.27\%&42.1\%&29.93\%&58.59\%&14.29\%&\textbf{61.62}\%\\
carpet&45.5\%&40.27\%&38.84\%&26.53\%&\textbf{50.86}\%&9.09\%&50.0\%\\
castle&45.85\%&44.28\%&40.36\%&32.98\%&47.93\%&16.67\%&\textbf{48.76}\%\\
 cat&68.78\%&45.44\%&\textbf{79.12}\%&54.63\%&75.53\%&25.0\%&73.4\%\\
cave&43.14\%&38.85\%&32.73\%&32.59\%&56.3\%&14.29\%&\textbf{57.98}\%\\
ceiling&91.88\%&90.6\%&90.15\%&85.66\%&\textbf{93.33}\%&87.61\%&93.33\%\\
ceramic&45.0\%&45.93\%&39.56\%&37.61\%&\textbf{61.03}\%&14.29\%&58.09\%\\
chains&72.22\%&83.33\%&77.78\%&66.67\%&\textbf{90.62}\%&33.33\%&87.5\%\\
chair&40.93\%&34.39\%&48.33\%&32.91\%&\textbf{54.44}\%&12.5\%&52.07\%\\
cheese&44.74\%&25.8\%&42.22\%&23.2\%&\textbf{49.32}\%&10.0\%&47.49\%\\
chicken&46.29\%&21.88\%&47.33\%&34.19\%&51.61\%&10.0\%&\textbf{55.91}\%\\
chocolate&58.92\%&48.5\%&52.73\%&38.92\%&\textbf{72.58}\%&16.67\%&69.35\%\\
church&58.59\%&57.7\%&39.48\%&25.27\%&57.36\%&12.5\%&\textbf{60.47}\%\\
city&55.88\%&24.37\%&65.79\%&30.03\%&\textbf{73.56}\%&14.29\%&72.41\%\\
clay&48.51\%&45.14\%&48.94\%&25.0\%&67.27\%&12.5\%&\textbf{69.09}\%\\
cliff&\textbf{75.69}\%&44.12\%&49.31\%&39.31\%&61.04\%&20.0\%&68.83\%\\
clock&57.49\%&44.98\%&45.37\%&26.63\%&\textbf{59.52}\%&12.5\%&57.94\%\\
clothes&42.63\%&25.6\%&37.46\%&28.6\%&62.5\%&12.5\%&\textbf{67.71}\%\\
cloud&23.97\%&16.67\%&19.89\%&23.89\%&44.0\%&12.5\%&\textbf{47.0}\%\\
coal&59.11\%&44.0\%&35.11\%&42.25\%&75.0\%&20.0\%&\textbf{77.78}\%\\
coast&59.92\%&42.47\%&54.6\%&44.51\%&66.67\%&25.0\%&\textbf{66.67}\%\\
coat&40.34\%&34.69\%&44.43\%&18.03\%&\textbf{51.38}\%&11.11\%&44.75\%\\
coffee&62.81\%&43.82\%&50.75\%&46.43\%&\textbf{76.58}\%&16.67\%&72.97\%\\
coin&66.47\%&59.05\%&63.84\%&33.26\%&71.53\%&18.67\%&\textbf{72.26}\%\\
column&\textbf{100.0}\%&100.0\%&100.0\%&100.0\%&100.0\%&100.0\%&100.0\%\\
computer&65.89\%&62.14\%&63.57\%&51.43\%&61.9\%&20.0\%&\textbf{73.02}\%\\
concrete&61.0\%&50.06\%&52.38\%&40.38\%&\textbf{69.44}\%&14.29\%&68.75\%\\
cookie&38.86\%&25.16\%&38.09\%&29.53\%&56.98\%&11.11\%&\textbf{59.3}\%\\
copper&49.41\%&40.39\%&44.07\%&33.18\%&\textbf{67.55}\%&14.29\%&61.59\%\\
cord&90.0\%&90.0\%&\textbf{100.0}\%&100.0\%&85.71\%&90.0\%&85.71\%\\
cotton&49.91\%&46.08\%&41.8\%&24.37\%&\textbf{54.01}\%&9.09\%&52.32\%\\
creek&67.22\%&66.22\%&65.89\%&61.06\%&82.19\%&20.0\%&\textbf{83.56}\%\\
deck&56.25\%&56.25\%&53.12\%&65.62\%&76.32\%&50.0\%&\textbf{86.84}\%\\
desert&51.25\%&40.04\%&45.4\%&35.39\%&\textbf{67.24}\%&16.67\%&62.93\%\\
desk&46.54\%&44.87\%&45.94\%&28.57\%&\textbf{54.01}\%&14.29\%&49.64\%\\
diamond&66.67\%&66.67\%&57.64\%&59.72\%&76.32\%&33.33\%&\textbf{76.32}\%\\
dirt&47.14\%&27.34\%&34.61\%&34.22\%&\textbf{52.88}\%&14.29\%&47.12\%\\
dog&67.1\%&43.99\%&58.09\%&45.27\%&66.67\%&11.11\%&\textbf{73.46}\%\\
door&61.67\%&53.57\%&49.64\%&31.07\%&\textbf{70.27}\%&20.0\%&68.92\%\\
dress&40.33\%&30.93\%&34.56\%&26.39\%&\textbf{52.7}\%&14.29\%&52.7\%\\
 \hline
\end{tabular}

\begin{tabular}{ |p{1.5cm}|p{1.5cm}|p{1.5cm}|p{1.5cm}|p{1.5cm}|p{1.5cm}|p{1.5cm}|p{1.5cm}|}
 \hline
 \multicolumn{8}{|c|}{Results for Metrics on Noun-Adjective Pairs} \\
 \hline
 Noun & Cosine Similarity & Dot Product & Binary Index & Product Quantiza- tion & Naive Bayes & Custom Neural Network & Basis Vectors\\
 \hline
 drum&\textbf{100.0}\%&100.0\%&100.0\%&100.0\%&100.0\%&100.0\%&100.0\%\\
dust&76.19\%&61.9\%&71.43\%&71.43\%&\textbf{82.14}\%&78.57\%&82.14\%\\
eggs&42.15\%&25.8\%&44.21\%&23.93\%&50.61\%&10.0\%&\textbf{56.1}\%\\
elephant&30.79\%&31.77\%&\textbf{44.88}\%&32.28\%&22.77\%&14.29\%&31.68\%\\
envelope&\textbf{74.39}\%&73.74\%&52.22\%&29.52\%&58.93\%&16.67\%&69.64\%\\
fabric&53.69\%&47.04\%&53.66\%&32.09\%&\textbf{70.97}\%&11.11\%&70.51\%\\
fan&57.97\%&48.96\%&37.29\%&36.16\%&66.67\%&16.67\%&\textbf{73.08}\%\\
farm&44.22\%&37.53\%&47.37\%&34.17\%&\textbf{56.41}\%&10.0\%&54.49\%\\
fence&55.11\%&50.99\%&53.98\%&34.27\%&\textbf{69.85}\%&20.41\%&66.91\%\\
field&48.68\%&43.79\%&43.15\%&28.03\%&\textbf{60.44}\%&20.0\%&51.65\%\\
fig&53.33\%&44.07\%&46.32\%&35.53\%&54.32\%&23.86\%&\textbf{58.02}\%\\
fire&\textbf{100.0}\%&100.0\%&82.86\%&82.86\%&100.0\%&50.0\%&100.0\%\\
fish&59.49\%&53.99\%&54.84\%&26.8\%&65.91\%&10.0\%&\textbf{65.91}\%\\
flame&\textbf{100.0}\%&100.0\%&100.0\%&100.0\%&100.0\%&100.0\%&100.0\%\\
 floor&58.17\%&48.82\%&62.1\%&45.91\%&67.14\%&25.0\%&\textbf{71.43}\%\\
flower&76.95\%&47.82\%&76.92\%&76.37\%&\textbf{86.96}\%&25.0\%&83.7\%\\
foam&52.99\%&52.22\%&41.13\%&26.97\%&\textbf{62.96}\%&12.5\%&62.96\%\\
forest&39.35\%&21.31\%&36.42\%&33.58\%&56.21\%&11.11\%&\textbf{56.8}\%\\
frame&62.05\%&45.46\%&68.97\%&70.54\%&\textbf{75.64}\%&25.0\%&74.36\%\\
fruit&42.63\%&17.19\%&37.35\%&32.6\%&\textbf{55.96}\%&10.0\%&52.75\%\\
furniture&56.1\%&45.54\%&38.3\%&36.12\%&74.47\%&14.29\%&\textbf{76.6}\%\\
garage&52.98\%&42.27\%&45.64\%&36.42\%&55.34\%&14.29\%&\textbf{63.11}\%\\
garden&63.49\%&65.28\%&73.21\%&67.26\%&72.73\%&33.33\%&\textbf{74.24}\%\\
garlic&75.0\%&33.33\%&72.22\%&75.35\%&81.82\%&25.0\%&\textbf{86.36}\%\\
gate&53.35\%&61.98\%&54.46\%&39.56\%&65.28\%&20.0\%&\textbf{70.83}\%\\
gear&64.92\%&48.56\%&64.87\%&32.29\%&\textbf{76.79}\%&20.0\%&73.21\%\\
gemstone&59.35\%&50.26\%&46.42\%&63.43\%&\textbf{71.67}\%&33.33\%&68.33\%\\
glass&38.93\%&39.0\%&43.31\%&25.05\%&59.5\%&11.11\%&\textbf{62.81}\%\\
glasses&87.5\%&84.38\%&90.62\%&88.64\%&92.11\%&50.0\%&\textbf{92.11}\%\\
granite&36.02\%&39.85\%&47.74\%&22.31\%&\textbf{59.88}\%&11.11\%&53.29\%\\
ground&46.88\%&40.51\%&45.7\%&36.1\%&\textbf{66.92}\%&12.5\%&63.08\%\\
handle&48.45\%&49.59\%&39.24\%&29.73\%&50.41\%&12.5\%&\textbf{53.66}\%\\
hat&85.22\%&63.81\%&73.79\%&\textbf{90.87}\%&89.36\%&33.33\%&87.23\%\\
highway&21.75\%&14.24\%&26.11\%&22.74\%&\textbf{43.55}\%&10.0\%&39.52\%\\
horse&65.27\%&40.39\%&57.09\%&45.59\%&\textbf{74.42}\%&20.0\%&68.6\%\\
hose&63.83\%&45.83\%&58.83\%&41.06\%&77.36\%&25.0\%&\textbf{77.36}\%\\
house&52.1\%&46.29\%&38.6\%&30.97\%&\textbf{68.37}\%&10.0\%&65.31\%\\
ice&46.28\%&24.97\%&35.77\%&30.04\%&50.76\%&11.11\%&\textbf{52.27}\%\\
iguana&34.16\%&26.32\%&31.77\%&27.04\%&\textbf{41.73}\%&16.67\%&38.58\%\\
island&49.92\%&40.82\%&49.12\%&41.08\%&56.0\%&16.67\%&\textbf{63.0}\%\\
jacket&54.73\%&52.67\%&55.44\%&35.9\%&\textbf{65.89}\%&16.67\%&62.79\%\\
jewelry&67.13\%&53.67\%&63.33\%&25.14\%&71.43\%&20.0\%&\textbf{72.32}\%\\
jungle&46.56\%&25.0\%&40.31\%&51.01\%&\textbf{71.15}\%&25.0\%&69.23\%\\
key&\textbf{93.45}\%&90.48\%&80.36\%&50.0\%&87.88\%&50.0\%&90.91\%\\
keyboard&72.84\%&66.05\%&69.44\%&61.11\%&68.75\%&33.33\%&\textbf{87.5}\%\\
kitchen&38.3\%&30.67\%&34.44\%&29.19\%&45.95\%&14.29\%&\textbf{45.95}\%\\
knife&38.39\%&35.84\%&32.93\%&14.51\%&\textbf{52.5}\%&8.33\%&49.0\%\\
lake&41.41\%&39.42\%&35.17\%&26.1\%&47.57\%&10.0\%&\textbf{49.19}\%\\
laptop&42.74\%&34.41\%&50.8\%&38.55\%&57.97\%&20.0\%&\textbf{63.77}\%\\
lead&55.83\%&37.58\%&\textbf{65.33}\%&45.83\%&52.0\%&20.83\%&54.0\%\\
leaf&49.85\%&29.43\%&49.43\%&28.24\%&\textbf{69.31}\%&14.29\%&67.33\%\\
 \hline
\end{tabular}

\begin{tabular}{ |p{1.5cm}|p{1.5cm}|p{1.5cm}|p{1.5cm}|p{1.5cm}|p{1.5cm}|p{1.5cm}|p{1.5cm}|}
 \hline
 \multicolumn{8}{|c|}{Results for Metrics on Noun-Adjective Pairs} \\
 \hline
 Noun & Cosine Similarity & Dot Product & Binary Index & Product Quantiza- tion & Naive Bayes & Custom Neural Network & Basis Vectors\\
 \hline
 lemon&44.64\%&25.34\%&50.59\%&35.62\%&\textbf{56.72}\%&16.67\%&52.24\%\\
library&36.38\%&26.67\%&29.13\%&36.56\%&\textbf{51.92}\%&12.5\%&51.92\%\\
lightbulb&25.18\%&13.74\%&27.28\%&21.17\%&\textbf{39.67}\%&9.09\%&33.88\%\\
lightning&50.0\%&50.0\%&50.0\%&66.67\%&\textbf{77.14}\%&50.0\%&74.29\%\\
log&42.22\%&37.3\%&52.3\%&35.83\%&57.89\%&14.29\%&\textbf{60.53}\%\\
mat&71.98\%&64.2\%&70.75\%&64.2\%&\textbf{85.71}\%&33.33\%&82.86\%\\
meat&21.71\%&15.35\%&26.57\%&17.55\%&\textbf{39.09}\%&8.33\%&34.01\%\\
metal&57.19\%&35.82\%&49.27\%&43.32\%&\textbf{74.59}\%&12.5\%&68.65\%\\
milk&50.0\%&35.67\%&37.33\%&39.7\%&75.68\%&20.0\%&\textbf{75.68}\%\\
mirror&50.53\%&44.14\%&43.75\%&30.87\%&\textbf{66.67}\%&20.0\%&61.9\%\\
moss&40.92\%&25.64\%&29.22\%&33.96\%&\textbf{61.22}\%&16.67\%&55.1\%\\
mountain&\textbf{72.23}\%&50.79\%&44.6\%&32.01\%&64.13\%&16.67\%&66.3\%\\
mud&44.67\%&28.82\%&34.16\%&24.04\%&\textbf{58.08}\%&9.09\%&55.56\%\\
necklace&35.91\%&27.79\%&34.37\%&29.51\%&53.88\%&11.11\%&\textbf{54.37}\%\\
 nest&68.77\%&56.79\%&60.08\%&55.19\%&71.67\%&33.33\%&\textbf{81.67}\%\\
 newspaper&54.2\%&55.97\%&53.85\%&39.06\%&64.44\%&41.67\%&\textbf{68.89}\%\\
nut&62.92\%&56.14\%&56.96\%&38.96\%&65.28\%&20.0\%&\textbf{73.61}\%\\
ocean&70.24\%&40.48\%&42.6\%&51.2\%&64.29\%&16.67\%&\textbf{75.71}\%\\
oil&\textbf{100.0}\%&100.0\%&100.0\%&75.0\%&83.33\%&100.0\%&100.0\%\\
orange&\textbf{79.9}\%&53.53\%&65.88\%&58.33\%&64.1\%&20.0\%&69.23\%\\
paint&\textbf{57.8}\%&52.95\%&45.22\%&26.96\%&50.96\%&14.29\%&50.96\%\\
palm&50.0\%&50.0\%&50.0\%&88.0\%&88.46\%&96.0\%&\textbf{96.15}\%\\
pants&60.33\%&45.67\%&48.24\%&19.21\%&58.62\%&9.09\%&\textbf{63.22}\%\\
paper&47.19\%&22.87\%&46.98\%&37.19\%&65.88\%&14.29\%&\textbf{67.06}\%\\
pasta&42.0\%&30.12\%&40.31\%&26.86\%&61.82\%&14.29\%&\textbf{63.64}\%\\
paste&39.82\%&30.81\%&37.36\%&30.1\%&40.74\%&12.5\%&\textbf{41.8}\%\\
pear&28.95\%&26.76\%&31.44\%&20.85\%&\textbf{41.67}\%&11.11\%&40.74\%\\
penny&\textbf{100.0}\%&85.71\%&78.57\%&78.57\%&100.0\%&100.0\%&100.0\%\\
persimmon&45.77\%&44.23\%&32.58\%&24.12\%&65.96\%&20.0\%&\textbf{65.96}\%\\
phone&56.13\%&36.58\%&52.35\%&41.94\%&65.18\%&14.29\%&\textbf{69.64}\%\\
pie&43.13\%&17.57\%&36.34\%&20.51\%&\textbf{46.96}\%&12.5\%&43.48\%\\
pizza&39.58\%&19.89\%&40.96\%&27.3\%&49.23\%&11.11\%&\textbf{50.77}\%\\
plant&38.59\%&22.24\%&42.5\%&27.02\%&47.56\%&10.0\%&\textbf{54.88}\%\\
plastic&55.79\%&41.22\%&49.02\%&24.29\%&70.75\%&14.29\%&\textbf{72.79}\%\\
plate&81.57\%&81.82\%&76.26\%&83.33\%&\textbf{91.89}\%&50.0\%&89.19\%\\
pond&41.0\%&26.34\%&38.15\%&28.66\%&43.95\%&12.5\%&\textbf{47.13}\%\\
pool&46.41\%&40.3\%&55.49\%&44.51\%&57.48\%&12.5\%&\textbf{61.42}\%\\
pot&83.24\%&78.39\%&68.09\%&45.19\%&88.16\%&20.0\%&\textbf{89.47}\%\\
potato&48.47\%&39.31\%&46.02\%&23.2\%&54.49\%&11.11\%&\textbf{55.09}\%\\
redwood&53.28\%&30.58\%&47.08\%&44.55\%&59.56\%&14.29\%&\textbf{61.03}\%\\
ribbon&\textbf{52.69}\%&30.28\%&37.52\%&20.87\%&46.86\%&10.0\%&49.14\%\\
ring&\textbf{55.6}\%&52.56\%&47.91\%&25.17\%&46.99\%&12.5\%&48.19\%\\
river&69.16\%&52.68\%&63.84\%&50.05\%&\textbf{76.47}\%&16.67\%&71.57\%\\
road&45.35\%&27.9\%&42.47\%&22.73\%&\textbf{50.32}\%&11.11\%&48.41\%\\
rock&51.89\%&51.87\%&51.73\%&36.26\%&\textbf{63.37}\%&12.5\%&57.56\%\\
roof&70.46\%&64.39\%&68.12\%&53.36\%&82.56\%&33.37\%&\textbf{84.88}\%\\
room&53.01\%&46.21\%&54.53\%&30.99\%&57.79\%&9.09\%&\textbf{57.79}\%\\
roots&48.89\%&42.22\%&53.33\%&70.83\%&60.0\%&33.33\%&\textbf{90.0}\%\\
rope&45.62\%&34.95\%&41.54\%&42.88\%&51.45\%&12.5\%&\textbf{61.59}\%\\
rubber&52.6\%&47.72\%&50.35\%&39.41\%&59.02\%&16.07\%&\textbf{59.84}\%\\
 \hline
\end{tabular}

\begin{tabular}{ |p{1.5cm}|p{1.5cm}|p{1.5cm}|p{1.5cm}|p{1.5cm}|p{1.5cm}|p{1.5cm}|p{1.5cm}|}
 \hline
 \multicolumn{8}{|c|}{Results for Metrics on Noun-Adjective Pairs} \\
 \hline
 Noun & Cosine Similarity & Dot Product & Binary Index & Product Quantiza- tion & Naive Bayes & Custom Neural Network & Basis Vectors\\
 \hline
 salad&38.98\%&17.9\%&36.94\%&39.97\%&57.86\%&16.67\%&\textbf{64.29}\%\\
salmon&39.89\%&33.84\%&42.89\%&40.32\%&\textbf{61.15}\%&14.29\%&51.8\%\\
sand&45.94\%&27.73\%&35.54\%&37.42\%&49.59\%&12.5\%&\textbf{56.1}\%\\
sandwich&56.29\%&36.04\%&51.28\%&37.29\%&55.38\%&16.67\%&\textbf{64.62}\%\\
sauce&54.32\%&47.36\%&55.15\%&40.0\%&57.69\%&20.0\%&\textbf{65.38}\%\\
screw&\textbf{98.08}\%&95.9\%&76.42\%&70.9\%&97.96\%&98.08\%&95.92\%\\
sea&88.1\%&88.1\%&88.1\%&83.33\%&\textbf{91.3}\%&50.0\%&82.61\%\\
seafood&35.49\%&17.86\%&29.58\%&32.09\%&\textbf{49.29}\%&14.29\%&45.71\%\\
shell&64.81\%&57.53\%&62.51\%&50.3\%&81.13\%&20.0\%&\textbf{84.91}\%\\
shirt&48.32\%&43.62\%&42.61\%&27.6\%&\textbf{57.21}\%&10.0\%&56.72\%\\
shoes&49.8\%&45.9\%&46.62\%&36.96\%&58.51\%&20.0\%&\textbf{62.77}\%\\
shore&36.92\%&30.85\%&51.1\%&33.58\%&\textbf{65.71}\%&16.67\%&62.86\%\\
shorts&46.51\%&46.13\%&32.35\%&21.19\%&52.52\%&11.11\%&\textbf{53.24}\%\\
shower&58.65\%&43.91\%&75.0\%&75.0\%&77.27\%&25.0\%&\textbf{81.82}\%\\
silk&42.98\%&40.39\%&39.95\%&22.13\%&\textbf{50.95}\%&11.11\%&50.95\%\\
sky&26.36\%&20.7\%&28.33\%&18.43\%&\textbf{40.0}\%&11.11\%&39.38\%\\
smoke&62.34\%&35.71\%&56.49\%&47.62\%&84.38\%&33.33\%&\textbf{84.38}\%\\
snake&55.06\%&44.32\%&49.43\%&39.49\%&60.22\%&16.67\%&\textbf{63.44}\%\\
snow&27.7\%&16.67\%&28.2\%&28.81\%&52.94\%&16.67\%&\textbf{56.47}\%\\
soup&28.25\%&28.38\%&32.0\%&30.75\%&48.28\%&11.11\%&\textbf{51.72}\%\\
steel&65.66\%&38.5\%&51.45\%&37.93\%&71.43\%&12.5\%&\textbf{71.43}\%\\
steps&67.52\%&52.38\%&73.46\%&50.54\%&\textbf{85.0}\%&20.0\%&80.0\%\\
stone&58.51\%&45.08\%&60.81\%&43.63\%&\textbf{75.95}\%&14.29\%&72.15\%\\
stream&53.01\%&44.25\%&51.48\%&31.19\%&\textbf{58.44}\%&14.29\%&58.44\%\\
street&56.32\%&36.45\%&54.5\%&27.25\%&59.78\%&10.0\%&\textbf{60.34}\%\\
sugar&45.42\%&28.5\%&39.89\%&26.91\%&\textbf{52.48}\%&14.29\%&49.5\%\\
sword&32.57\%&32.61\%&36.18\%&28.84\%&49.32\%&16.67\%&\textbf{54.79}\%\\
table&\textbf{100.0}\%&100.0\%&100.0\%&100.0\%&100.0\%&100.0\%&100.0\%\\
tea&50.58\%&35.9\%&70.11\%&40.51\%&65.96\%&33.33\%&\textbf{76.6}\%\\
thread&46.67\%&46.67\%&38.33\%&48.33\%&88.89\%&33.33\%&\textbf{88.89}\%\\
tie&83.33\%&83.33\%&58.33\%&58.33\%&93.55\%&83.33\%&\textbf{96.77}\%\\
tiger&33.53\%&25.0\%&31.77\%&37.41\%&\textbf{50.0}\%&25.0\%&48.28\%\\
tile&40.35\%&19.72\%&32.8\%&26.97\%&50.68\%&11.11\%&\textbf{52.74}\%\\
tire&51.7\%&47.85\%&52.88\%&30.82\%&69.07\%&16.67\%&\textbf{75.26}\%\\
tomato&40.11\%&13.46\%&28.08\%&17.87\%&\textbf{56.0}\%&8.33\%&51.33\%\\
tower&58.7\%&51.88\%&63.92\%&49.5\%&67.47\%&14.29\%&\textbf{68.67}\%\\
town&64.84\%&65.56\%&75.25\%&58.1\%&\textbf{79.01}\%&20.0\%&76.54\%\\
toy&41.76\%&26.47\%&37.55\%&24.35\%&\textbf{59.21}\%&10.0\%&58.55\%\\
trail&33.85\%&26.63\%&31.68\%&24.83\%&47.67\%&20.0\%&\textbf{48.84}\%\\
tree&65.5\%&45.54\%&64.84\%&30.41\%&\textbf{71.28}\%&10.0\%&71.28\%\\
truck&50.26\%&44.98\%&52.47\%&34.75\%&54.01\%&11.11\%&\textbf{59.36}\%\\
tube&66.95\%&64.31\%&58.98\%&43.07\%&70.27\%&14.29\%&\textbf{71.17}\%\\
tulip&76.72\%&50.0\%&77.25\%&83.6\%&81.25\%&50.0\%&\textbf{89.58}\%\\
vacuum&\textbf{100.0}\%&100.0\%&100.0\%&100.0\%&100.0\%&100.0\%&100.0\%\\
valley&47.41\%&40.27\%&39.4\%&26.99\%&\textbf{54.19}\%&9.09\%&53.55\%\\
vegetable&47.02\%&26.08\%&45.06\%&27.5\%&\textbf{57.58}\%&12.5\%&56.36\%\\
velvet&33.23\%&25.76\%&42.72\%&22.4\%&\textbf{60.29}\%&12.5\%&55.15\%\\
wall&63.14\%&54.26\%&56.24\%&31.31\%&70.62\%&11.11\%&\textbf{72.5}\%\\
water&82.14\%&59.23\%&\textbf{86.31}\%&69.94\%&84.85\%&25.0\%&78.79\%\\
wave&44.91\%&23.2\%&35.09\%&33.57\%&55.91\%&20.0\%&\textbf{59.14}\%\\
wax&47.47\%&28.19\%&30.84\%&28.28\%&\textbf{50.75}\%&16.67\%&38.81\%\\
 \hline
\end{tabular}\\

\begin{tabular}{ |p{1.5cm}|p{1.5cm}|p{1.5cm}|p{1.5cm}|p{1.5cm}|p{1.5cm}|p{1.5cm}|p{1.5cm}|}
 \hline
 \multicolumn{8}{|c|}{Results for Metrics on Noun-Adjective Pairs} \\
 \hline
 Noun & Cosine Similarity & Dot Product & Binary Index & Product Quantiza- tion & Naive Bayes & Custom Neural Network & Basis Vectors\\
 \hline
 well&86.83\%&88.49\%&80.48\%&77.78\%&84.91\%&66.67\%&\textbf{88.68}\%\\
wheel&51.52\%&51.52\%&37.58\%&17.99\%&62.5\%&16.67\%&\textbf{64.29}\%\\
window&\textbf{74.66}\%&57.89\%&47.37\%&29.71\%&68.18\%&21.43\%&69.7\%\\
wire&75.62\%&62.95\%&65.9\%&48.54\%&\textbf{81.25}\%&20.0\%&79.69\%\\
wood&46.15\%&16.79\%&33.1\%&25.6\%&\textbf{59.68}\%&12.5\%&55.38\%\\
wool&36.57\%&34.83\%&33.94\%&22.2\%&\textbf{51.42}\%&10.0\%&47.77\%\\
 \hline
\end{tabular}\\

\end{center}

\newpage
Our full results for our adjectives experiment using CLIP-ViT-Large-Patch14: \\

\begin{center}
\begin{tabular}{ 
|p{1.5cm}|p{1.5cm}|p{1.5cm}|}
 \hline
 \multicolumn{3}{|c|}{Results for Metrics on Adjectives} \\
 \hline
 Adjective & Logistic Regression & Basis Vectors (BVM)\\
\hline
ancient&54.26\%&\textbf{64.67\%}\\
barren&40.0\%&\textbf{48.54\%}\\
bent&\textbf{37.5\%}&18.65\%\\
blunt&\textbf{18.18\%}&0.0\%\\
bright&\textbf{42.86\%}&23.99\%\\
broken&44.57\%&\textbf{47.76\%}\\
browned&\textbf{31.58\%}&14.09\%\\
brushed&59.26\%&\textbf{59.65\%}\\
burnt&\textbf{54.92\%}&52.45\%\\
caramelized&50.76\%&\textbf{76.07\%}\\
chipped&\textbf{29.07\%}&7.34\%\\
clean&\textbf{38.67\%}&23.0\%\\
clear&\textbf{46.94\%}&28.95\%\\
closed&\textbf{26.67\%}&0.0\%\\
cloudy&\textbf{50.0\%}&0.0\%\\
cluttered&60.87\%&\textbf{69.0\%}\\
coiled&66.85\%&\textbf{75.55\%}\\
cooked&38.27\%&\textbf{42.01\%}\\
cored&\textbf{32.61\%}&2.65\%\\
cracked&\textbf{50.57\%}&47.19\%\\
creased&\textbf{28.95\%}&6.16\%\\
crinkled&\textbf{34.11\%}&15.65\%\\
crumpled&\textbf{29.82\%}&24.92\%\\
crushed&\textbf{41.36\%}&37.89\%\\
curved&\textbf{37.36\%}&23.08\%\\
cut&\textbf{36.28\%}&22.12\%\\
damp&\textbf{18.64\%}&3.39\%\\
dark&\textbf{62.5\%}&60.94\%\\
deflated&\textbf{59.46\%}&18.92\%\\
dented&\textbf{31.48\%}&9.26\%\\
diced&31.58\%&\textbf{34.21\%}\\
dirty&\textbf{42.62\%}&22.95\%\\
draped&61.46\%&\textbf{69.79\%}\\
dry&\textbf{48.24\%}&35.29\%\\
dull&\textbf{23.81\%}&0.0\%\\
empty&\textbf{43.27\%}&\textbf{43.27\%}\\
engraved&64.71\%&\textbf{84.87\%}\\

\hline
\end{tabular}

\begin{tabular}{ 
|p{1.5cm}|p{1.5cm}|p{1.5cm}|}
 \hline
 \multicolumn{3}{|c|}{Results for Metrics on Adjectives} \\
 \hline
 Adjective & Logistic Regression & Basis Vectors (BVM)\\
\hline
eroded&58.04\%&\textbf{72.03\%}\\
fallen&\textbf{0.0\%}&\textbf{0.0\%}\\
filled&\textbf{21.88\%}&3.12\%\\
foggy&70.0\%&\textbf{87.5\%}\\
folded&\textbf{48.61\%}&47.22\%\\
frayed&\textbf{50.49\%}&25.24\%\\
fresh&\textbf{48.68\%}&45.39\%\\
frozen&\textbf{50.91\%}&50.0\%\\
full&\textbf{40.0\%}&0.0\%\\
grimy&\textbf{3.33\%}&0.0\%\\
heavy&\textbf{36.54\%}&17.31\%\\
huge&42.75\%&\textbf{48.55\%}\\
inflated&50.94\%&\textbf{69.81\%}\\
large&\textbf{28.78\%}&20.86\%\\
lightweight&\textbf{26.67\%}&25.0\%\\
loose&\textbf{28.57\%}&8.57\%\\
mashed&\textbf{43.75\%}&25.0\%\\
melted&\textbf{34.38\%}&18.75\%\\
modern&\textbf{52.81\%}&\textbf{52.81\%}\\
moldy&44.62\%&\textbf{53.85\%}\\
molten&\textbf{51.75\%}&44.74\%\\
mossy&66.84\%&\textbf{76.47\%}\\
muddy&\textbf{56.38\%}&45.74\%\\
murky&\textbf{29.41\%}&5.88\%\\
narrow&\textbf{24.42\%}&1.16\%\\
new&43.92\%&\textbf{55.41\%}\\
old&\textbf{54.91\%}&53.57\%\\
open&\textbf{66.67\%}&5.56\%\\
painted&\textbf{50.0\%}&48.15\%\\
peeled&\textbf{47.06\%}&\textbf{47.06\%}\\
pierced&\textbf{56.86\%}&37.25\%\\
pressed&\textbf{36.79\%}&13.21\%\\
pureed&40.4\%&\textbf{45.45\%}\\
raw&59.76\%&\textbf{63.41\%}\\
ripe&62.5\%&\textbf{67.71\%}\\
ripped&\textbf{56.0\%}&52.0\%\\
rough&20.0\%&\textbf{26.67\%}\\
ruffled&66.67\%&\textbf{77.27\%}\\
runny&\textbf{0.0\%}&\textbf{0.0\%}\\
rusty&61.4\%&\textbf{70.76\%}\\
scratched&\textbf{44.64\%}&23.21\%\\
sharp&\textbf{38.89\%}&22.22\%\\
shattered&\textbf{38.18\%}&0.0\%\\
shiny&\textbf{23.08\%}&0.0\%\\
short&\textbf{58.33\%}&16.67\%\\
sliced&44.36\%&\textbf{45.11\%}\\
small&\textbf{39.06\%}&36.25\%\\
smooth&\textbf{18.75\%}&0.0\%\\
spilled&\textbf{53.57\%}&42.86\%\\
\hline
\end{tabular}

\begin{tabular}{ 
|p{1.5cm}|p{1.5cm}|p{1.5cm}|}
 \hline
 \multicolumn{3}{|c|}{Results for Metrics on Adjectives} \\
 \hline
 Adjective & Logistic Regression & Basis Vectors (BVM)\\
\hline
splintered&\textbf{47.62\%}&0.0\%\\
squished&\textbf{16.67\%}&0.0\%\\
standing&\textbf{75.0\%}&0.0\%\\
steaming&\textbf{53.7\%}&50.93\%\\
straight&\textbf{64.0\%}&45.33\%\\
sunny&\textbf{41.82\%}&36.36\%\\
tall&\textbf{62.5\%}&50.0\%\\
thawed&\textbf{33.73\%}&26.51\%\\
thick&\textbf{25.89\%}&16.24\%\\
thin&\textbf{34.1\%}&24.86\%\\
tight&73.53\%&\textbf{76.47\%}\\
tiny&\textbf{38.74\%}&27.93\%\\
toppled&\textbf{35.29\%}&17.65\%\\
torn&\textbf{40.35\%}&19.3\%\\
unpainted&\textbf{33.33\%}&23.33\%\\
unripe&\textbf{55.38\%}&49.23\%\\
upright&\textbf{17.65\%}&0.0\%\\
verdant&\textbf{21.79\%}&15.38\%\\
viscous&\textbf{0.0\%}&\textbf{0.0\%}\\
weathered&44.03\%&\textbf{51.03\%}\\
wet&\textbf{30.77\%}&16.35\%\\
whipped&\textbf{29.51\%}&18.03\%\\
wide&\textbf{26.44\%}&16.09\%\\
wilted&\textbf{62.5\%}&25.0\%\\
windblown&\textbf{50.79\%}&39.68\%\\
winding&\textbf{35.14\%}&21.62\%\\
worn&\textbf{65.52\%}&31.03\%\\
wrinkled&\textbf{42.72\%}&20.39\%\\
young&52.17\%&\textbf{71.74\%}\\
\hline
\end{tabular}

\end{center}

\end{document}